\documentclass[
]{ceurart}

\usepackage{listings}
\begin{document}

\copyrightyear{2025}
\copyrightclause{Copyright for this paper by its authors.
  Use permitted under Creative Commons License Attribution 4.0
  International (CC BY 4.0).}

\conference{Joint proceedings of KBC-LM and LM-KBC @ ISWC 2025}

\title{Noise or Nuance: An Investigation Into Useful Information and Filtering For LLM Driven AKBC}

\author[1]{Alex Clay}[%
orcid= 0009-0004-2237-7412,
email=alex.clay@city.ac.uk,
%url=https://yamadharma.github.io/,
]
\cormark[1]
\author[1]{Ernesto Jiménez-Ruiz}[%
orcid= 0000-0002-9083-4599,
email=ernesto.jimenez-ruiz@city.ac.uk,
%url=https://ernestojimenezruiz.github.io,
]

\author[1,2]{Pranava Madhyastha}[
orcid = 0000-0002-4438-8161,
email = pranava.madhyastha@city.ac.uk,
]
\address[1]{City St George's, University of London,
Northampton Square, London, EC1V 0HB, United Kingdom}
\address[2]{The Alan Turing Institute, British Library, 96 Euston Rd., London NW1 2DB, United Kingdom}

\cortext[1]{Corresponding author.}
\begin{abstract}
RAG and fine-tuning are prevalent strategies for improving the quality of LLM outputs. However, in constrained situations, such as that of the 2025 LM-KBC challenge, such techniques are restricted. In this work we investigate three facets of the triple completion task: generation, quality assurance, and LLM response parsing. Our work finds that in this constrained setting: additional information improves generation quality, LLMs can be effective at filtering poor quality triples, and the tradeoff between flexibility and consistency with LLM response parsing is setting dependent. %is the LLM able to generate improved outputs without either of these techniques? We investigate the value of different supplemental information on automated triple completion, as well as means for filtering and extracting the candidate tail entities relying solely on the LLM and in-built python capabilities. 
\end{abstract}
\maketitle

\section{Introduction}
Retrieval Augmented Generation is often implemented to improve the performance of LLMs on a variety of tasks, including those pertaining to knowledge extraction ~\cite{saad-falcon-etal-2024-ares, 10.1145/3624918.3625336}. Automated Knowledge Base Completion (AKBC) has long been a topic of interest ~\cite{DBLP:conf/kbclm/KaloNRZ24}, and has more recently seen an increase in investigations relating to the use of LLMs, not only to create KBs ~\cite{hu-etal-2025-enabling, cohen-etal-2023-crawling}, but in some instances to act in place of them. We adopt one such setting and task, introduced by the 2025 LM-KBC challenge, to utilise only the given data and Qwen3's~\cite{yang2025qwen3technicalreport} 8 Billion parameter variation without finetuning or RAG to generate completions for KG triples\footnote{Our code: https://github.com/alexclay42/lmkbcWorkshop}. This provided the opportunity to investigate specific facets of what aspects of generation improve triple completion under such constraints. Therefore, our task was to generate high quality tail entities to complete a triple, without augmentation to the assigned LLM. We considered means of emulating aspects of RAG and finetuning without implementing either. 

The presence of grounding information when prompting LLMs can influence the quality of the generated output~\cite{tang-etal-2024-minicheck}. As such, even LLM generated supporting information might be able to improve efficacy of example-supported generation in situations with sparse external data. Moreover, without external data or a ground truth, evaluation of triple quality lacks a key point of comparison. LLMs have been utilised as judges~\cite{Hu_2025}, which provide a convenient means of checking generated outputs against the model's own knowledge. Additionally, handling and formatting of LLM responses are crucial to avoiding error propagation; while often convenient to use NLP libraries like SpaCy\footnote{https://spacy.io} for parsing, the LLM could potentially complete the task itself. In our work, we hypothesized that it is possible to utilise the LLM in a number of ways to holistically replace augmentation like finetuning and external knowledge sources through handling: grounding information for generation, filtration of poor quality triples, and cleaning of LLM responses. As such, we present an investigation of the following questions within the constrained setting of the LM-KBC challenge:\textit{
    \newline RQ1: What type of information is useful to generating candidate tails with limited access to external data?
    \newline RQ2: Without access to the ground truth, what is an effective means for filtering poor quality triples?
    \newline RQ3: Is the flexibility of LLM entity extraction worth the tradeoff of inconsistency?}
%We make our code and all experiments available \footnote{https://github.com/alexclay42/lmkbcWorkshop}.

Our primary findings indicate that within our constrained setting: additional information improves generation quality, LLMs are effective at filtering poor quality triples if not a bit strict, and the tradeoff between flexibility and consistency in LLM response parsing varies based on the setting.
\section{Related Work}
%- introduce the concepts/tasks
%Most knowledge bases focus on positive knowledge, or things that are known to be true https://arxiv.org/pdf/2305.05403. *use this to give better intro, these are the ppl who made the challeneg and workshiop so handle carefully*
In recent years, LLMs have been more commonly used for the task of automatic knowledge base completion. ~\citet{hu-etal-2025-enabling}, for instance, employed LLM-generated information about a given topic in the form of triples to retrieve new entities, applying the same technique to the newly returned entities. ~\citet{he2021empiricalstudyfewshotknowledge} meanwhile, focused on a setting with few available example triples, and determined that few-shot examples can improve one and two-hop relations. ~\citet{cohen-etal-2023-crawling} created a knowledge graph from a seed entity through the generation of potential relations and subsequent tails.  The creation of the knowledge graph relied entirely on the knowledge of the chosen LLM to provide candidates in this approach,  introducing paraphrasing of the entities during the relation generation and relations during tail generation to increase quantity and diversity of the candidates. In our work, we adapted the paraphrasing concept to be a brief factual explanation of the entity and used the same format as ~\cite{cohen-etal-2023-crawling} for the examples provided at generation. 

Masked Language models, such as BERT, have also been employed for this task. For example, in the original LAMA Probe, ~\citet{petroni-etal-2019-language} investigated BERT for answering fill-in-the-blank style queries. ~\citet{jiang-etal-2020-know} improved the SOTA on the LAMA task by over 8 percent by utilising more effective prompts through mining and paraphrasing-based methods. Additionally ~\citet{10.1007/978-3-031-33455-9_14} found that BERT had good generalization ability and that prompt formulation and vocabulary expansion could significantly improve its capability. Though in a different format, we apply a similar concept of eliciting the LLM to fill in the blank, of completing the given format rather than a sentence.

In order to derive understanding of LLMs as knowledge bases, many have investigated the information inherently possessed in an LLM's parameters. ~\citet{hu-etal-2025-enabling} noted that availability bias is a crucial issue as, essentially, only surface information is being reached, and that which is beyond the benchmarking or user-induced information is not being adequately explored. Moreover, LLMs can be sensitive to perturbations as the naming of entities can impact `model cognition'~\cite{yin-etal-2023-alcuna} and that LLMs are not terribly adept at relating new knowledge with knowledge they already possess. Moreover, ~\citet{berglund2024reversalcursellmstrained} found that LLMs typically do not learn to reverse relationships unless something like "A is B" appears in-context. When utilising LLMs, the information presented with and the prompt itself greatly influence the generated response. ~\citet{chen-etal-2023-mapo} indicated the importance of model-specific prompt adaptation, something avoided by \citet{Wu_2025}'s promptless knowledge estimator where they instead utilise similarly related entity and tail pairs. We utilise a similar promptless strategy in our approach, to more directly assess the impact of different information in eliciting generation of tail candidates.
\section{Methodology}
\begin{figure}[h]
    \centering
    \includegraphics[width=0.5\linewidth]{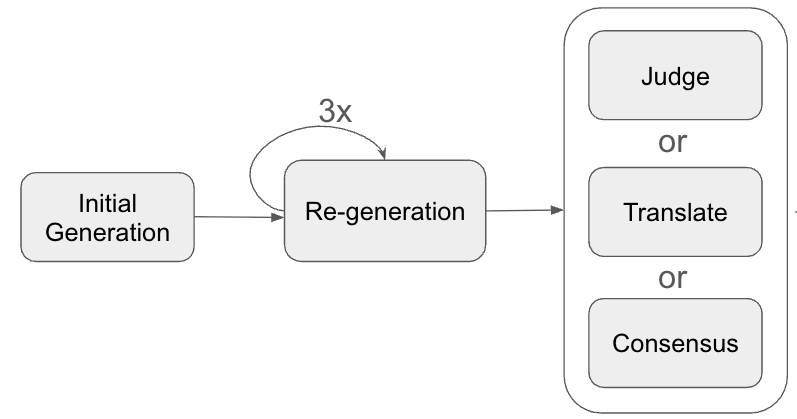}
    \caption{The component flow for the experiments. }
    \label{fig:placeholder}
\end{figure}

Our task, as with the LM-KBC challenge\footnote{https://lm-kbc.github.io/challenge2025/}, was to complete a set of given initial entity and relation pairs, and to elicit the designated LLM, Qwen3 8B, to provide a correct tail entity for each pair. A triple is the underlying structure in a knowledge graph with the format of two entities with a relation between them, the later of which is a 'tail' entity.  Some of the entity/relation pairs might have one, multiple, or no actual tail entity. Our research questions address three crucial aspects of this task generation, evaluation, and parsing; and how to successfully address it within the provided limitations. To address the task, we constructed a three step program: firstly an initial generation, then re-generation, and completing with a filtration process. We present an illustration in Figure \ref{fig:placeholder} showing the linear process and filtration variations.

We follow a general approach to generate an initial set of tail entities for the dataset, using the same example based prompt as ~\citet{cohen-etal-2023-crawling}. This initial set of candidates could then be sent to the re-generation component based on the experiment. The re-generation component was introduced to provide an opportunity for poor quality triples to be improved with the intention of reducing the number of candidates removed during filtering. Each triple was evaluated with a call to the LLM to judge its quality and regenerated on failure. This process could happen up to 3 times, if the candidate triple was continually rejected. The candidate triples then reached the filtration step, and either encountered the judge, consensus, or translation approach before formatting for evaluation. Following each component, the LLM response was cleaned into a candidate tail before proceeding (ie. 'The answer is Isla Nublar' becomes 'Isla Nublar'), a process which itself provided means for comparing cleaning techniques. Each of these components contributes to both the completion of the task and investigation of our research questions. 
%
%The filtration step provides a means for an investigation where the LLM is used to filter out poor triples, and evaluation of the extraction of candidate entities is carried out throughout the process at any point 
%
%We are able to explore different supplemental information as well as example types at two points, one in which the previous incorrect attempts are uniquely available. 
%\subsubsection{RQ1}
\paragraph{RQ1}In order to address the question of what information is useful to generation (RQ1), we modified the examples and additional information at the initial and re-generation points. Firstly, we investigated the introduction of LLM-generated supporting information about the target entity, gained through prompting the LLM to return facts about it. This information was provided in combination with the examples at initial generation and re-generation to determine if it could act similarly to RAG grounding in improving the efficacy. Another type of information we investigated was the addition of a guide sentence to indicate the expected output type (such as a name or country) in place of the LLM generated grounding. Many approaches, such as the baseline approach for the challenge\footnote{https://github.com/lm-kbc/dataset2025}, utilised relation-specific methods, in that each different type of relation had a specific type of prompt. We adapted this concept into this guide sentence to determine if it aided the LLM in producing the correct output. We also investigated variation in the example triples provided, comparing that of relation-specific and entity/relation pair specific examples. The relation specific examples were the same relation type as the target entity/relation pair, whereas the specific examples were selected on a pair by pair basis through using the LLM to select the three most similar examples of the same relation type from the LM-KBC validation dataset. The intention was to determine if having more similar examples might ground the generation in more similar information to that of the target, such as potentially returning examples with other islands if the target entity was an island. We aimed for the information grounding and specific examples to potentially act as a simulacra of RAG, and for the relation specific guides and previous errors to emulate fine tuning. 

%\subsubsection{RQ2}
\paragraph{RQ2}Due to the constrained setting, it was necessary to implement a means of quality control that did not rely on external verification, the focus of RQ2. Inspired by ~\citet{wiland-etal-2024-bear}, who used an approach which leveraged the LLM's ability to predict the log-likelihood of a statement to rank answer options, we utilised the LLM to indicate the quality of the triples. We compared three variations of filtration based off of different concepts: LLM as a judge ~\cite{Hu_2025}, consensus similar to the duplicate reduction and minimum instances in ~\citet{cohen-etal-2023-crawling}, and a comparison of translation between formats. For the judge filter, the LLM was prompted to return a value between 0 and 100 indicating the quality of the triple. This was done 3 times, and if the average score was 50 or more, the candidate triple was accepted. We use the same judge for the re-generation and filter, though increase the calls to 3 times and take the average, to better handle nonsensical answers and increase confidence of the ruling. The consensus filter required any predicted tail to appear twice or more to be accepted, as more frequent appearances might indicate higher confidence in a candidate tail. For the translation filter, the LLM was prompted to convert the candidate triple into a natural language sentence. The sentence and candidate triple were then provided in another LLM call to determine if they had the same meaning, with the intention of nonsensical candidate tails translating poorly into a sentence and therefore not matching the original meanings. 
%%We additionally explore whether the use of an LLM or python based consensus is more effective, which is discussed more thoroughly RQ3. 
%
%We added a judge, consensus, and a translation into natural language to investigate which of these means could effectively remove poor quality triples without access to a ground truth. 
%\subsubsection{RQ3}

\paragraph{RQ3}For question three, to understand the tradeoff between the flexibility and inconsistency of an LLM, we further conducted an ablation to compare the regex with LLM fallback we had been utilizing throughout to understand the degree to which the LLM was usefully handling what the regex could not. We similarly compared a configuration of the consensus without the LLM to equally understand if the LLM was successfully extracting and consolidating the candidate entities. 
%Throughout the process, it also became clear that the handling of LLM responses and the extraction of the relevant information therein was likely as important as some of the program components as error propagation can be encountered in situations with sequential generation steps~\cite{cohen-etal-2023-crawling}.  This contributed to the decision to create and evaluate a more structured version of both the cleaning functionality and the consensus components. While the cleaning function was created with an LLM fallback as default, the consensus was either comprised of LLM calls or the built-in python counter. Therefore additional investigations were carried out to understand the tradeoff between the more flexible LLM answers and more reliable but somewhat unforgiving regex handling. 
\section{Experiments and Results}
For our experiments we used a stratified subset of 100 samples from the LM-KBC 2025 training dataset, with examples extracted from the validation set. The sample was equivalent to about 1/5th of the entire dataset, and selected to due the long computation time of the re-generation component. The sample was randomly selected, maintaining a representative proportion for each relation type. For all LLM use, we employed Qwen3 8B through the transformers library \footnote{https://huggingface.co/Qwen/Qwen3-8B} without any fine tuning or RAG. The scores we present in the following portions are the values for F1, macro precision, and macro recall. For the purpose of this investigation, we take information to mean any additional data and examples supplied alongside a prompt to the LLM. Full details of our implementation \& prompts can be found in our code.
\subsection{RQ1: What type of information is useful to generating candidate tails with limited access to external data?}
For our first question, we began by investigating the impact of introducing LLM generated facts about the target entity. We found that  the introduction of such information was not useful to the re-generation component, as the performance decreased from the base version without information, as shown in Table \ref{tab:info}. As the primary difference between the initial generation and the re-generation was the use of examples instead of previous incorrect attempts, this is likely the reason that the initial generation improved with additional information while the re-generation did not. Therefore, the additional information may have acted primarily as noise when introduced with incorrect attempts but acted as a form of grounding with correct examples. 
\begin{table}[h]
\centering
\small
\begin{tabular}{lllll}
\toprule
\textbf{Approach} & \textbf{F1} & \textbf{Precision}  & \textbf{Recall}\\
\midrule
Base & .148 & .096 & .320 \\
Intial Generation & .151 & .097 & .338 \\
Re-generation & .136 & .085 & .340 \\
\bottomrule
\end{tabular}
\caption{Addition of information describing the target entity prior to examples at generation time.}
\label{tab:info}
\end{table}
In our next experiment, we evaluated the efficacy of positive and negative examples during the re-generation component. The re-generation component allows the unique opportunity to investigate whether the LLM is able to self-correct through the use of previous attempts. We find that within our setting, such information does improve precision more than examples, both of which display improvement over format-only elicitation as shown in Table \ref{tab:results}. Despite a lower precision, the correct examples yields a slightly higher F1, likely due to differing performance on certain relations, which is highlighted through the use of macro evaluation scoring. 
\begin{table}[h]
\centering
\small
\begin{tabular}{lllll}
\toprule
\textbf{Approach} & \textbf{F1} & \textbf{Precision}  & \textbf{Recall} \\
\midrule
Base & .126 & .078 & .325 \\
w/ Correct Examples & .151 & .097 & .338 \\
w/ Incorrect Examples (previous) & .158 & .103 & .338\\
\bottomrule
\end{tabular}
\caption{Re-generation experiments. The base version provides only the format without examples or additional information.}
\label{tab:results}
\end{table}
To determine the value of entity/relation pair specific examples and that of relation guides, we used a baseline of relation specific examples with additional LLM generated information. We compared this base with custom examples for the entity/relation pair, and a version where a relation guide was used in place of the background information. The relation guide provided a sentence stating the expected type, for instance, for \texttt{awardWonBy} it requested a list of names, and \texttt{countryLandBordersCountry} a list of countries. 

As anticipated, both the custom examples and relation guide improved on the original score as shown in Table \ref{tab:initialcomparison}. It is likely that as the examples were already specific to the relation type, there was reduced variance when introducing the specially selected examples. Had the base been relation unspecific it would likely have been possible to see a larger impact. The relation guide likely fulfilled its purpose in reducing some type irrelevant answers, which otherwise might have been difficult to achieve without additional processing.
\begin{table}[h]
\centering
\small
\begin{tabular}{lllll}
\toprule
\textbf{Approach} & \textbf{F1} & \textbf{Precision}  & \textbf{Recall}\\
\midrule
Relation Specific & .152 & .101 & .310 \\
Custom for Entity/Relation & .162 & .107 & .335 \\
With Relation Guide & .183 & .128 & .320 \\
\bottomrule
\end{tabular}
\caption{Comparison of additions to initial generation. Relation specific acts as the baseline.}
\label{tab:initialcomparison}
\end{table}
\subsection{RQ2: Without access to a ground truth, what is an effective means for filtering poor quality triples?} 
In question 2 we focus on the addition of a filter to the process to remove poor quality triples. The addition of any filter showed significant improvement in precision over the base version, as displayed in Table \ref{tab:judge}. The consistent temperature judge preformed best overall in F1, over that of the judges using an average of three different temperatures, possibly due to the reduced variance of the consistent judge and non-number or nonsensical responses from the differing temperature version. 
\begin{table}[h]
\centering
\small
\begin{tabular}{lllll}
\toprule
\textbf{Approach} & \textbf{F1} & \textbf{Precision}  & \textbf{Recall} \\
\midrule
Base (Initial only) & .152 & .101 & .310 \\
Judge & .360 & .438 & .305 \\
Judge Alt. Temps & .290 & .286 & .295 \\
Translate & .313 & .340 & .290 \\
Consensus & .205 & .147 & .341 \\
\bottomrule
\end{tabular}
\caption{Comparison of filtering approaches. The candidate triples were identical for each and immediately following the initial generation.}
\label{tab:judge}
\end{table}
Despite having good precision, the translation filter also had the lowest recall, implying that it may have been very strict in its reduction of answers, and highlighting a precision/recall tradeoff. Moreover, it involved two calls to the LLM, the first to convert the triple into a natural language sentence, and the second to compare the meaning of the two. This two step process may have introduced errors with propagated across, and decreased the number of retained candidates as a result. 

Interestingly, the recall is reduced as expected across all versions with the exception of the consensus. It would be expected that in filtering out candidates the recall would decrease, however due to the use of an LLM for the consensus, it is possible that it introduced new candidates during consolidation. 
\subsection{RQ3: Is the flexibility of LLM entity extraction worth the tradeoff of inconsistency?}
Finally, we conduct an ablation over the initial generation to determine the efficacy of the LLM in handling the entity extraction, and a comparison of the consensus using an LLM approach and a deterministic counter approach.
\begin{table}[h]
    \centering
    \begin{tabular}{lllll}
    \toprule
    \textbf{Approach} & \textbf{F1} & \textbf{Precision} & \textbf{Recall} \\
        \midrule
        Regex w/ LLM Fallback & .152 & .101 & .310\\
        Regex only & .283 & .268 &.300\\
        LLM only & .133 & .088 & .270 \\
        \bottomrule
    \end{tabular}
    \caption{Comparisons of the different processing of the LLM responses. We used the combination as default.}
    \label{tab:my_label}
\end{table}
In order to more thoroughly process the LLM responses, regular expression based extraction of numbers and capitalized sets of words, ideally proper nouns, initially was used with a fallback to an LLM if nothing was returned. We anticipated that the combination would be more effective than the regular expression alone, as the LLM might be able to extract additional valid answers. However, as the LLM only approach preformed poorly, and the version utilizing both also saw a reduction in score from the regex only version, as shown in Table \ref{tab:my_label} we hypothesize that the LLM's involvement may have contributed poor quality answers rather than simply catching differing format. Additionally, when constructing the implementation, we noticed that when the LLM generated a candidate: the phrasing, punctuation, and capitalization was often consistent each time it was generated. As such, the regex version's removal of such responses may not have actually been undesirable.
\begin{table}[h]
\centering
\small
\begin{tabular}{lllll}
\toprule
\textbf{Approach} & \textbf{F1} & \textbf{Precision}  & \textbf{Recall} \\
\midrule
LLM Consensus & .205 & .147 & .341\\
Regex Consensus & .147 & .098 & .290 \\
\bottomrule
\end{tabular}
\caption{Variations in consensus method.}
\label{tab:consensus}
\end{table}
When evaluating the difference between LLM and Python counter approaches for the consensus filter, we anticipated that due to the variability of responses, the LLM approach would yield a lower score. However as can be seen in Table \ref{tab:consensus}, the LLM actually yielded the highest scores. It is likely that this is due to the LLM being more flexible with it's answers and accommodating more values with the same meaning, whereas the Python counter function uses restrictive string matching. 

As such, the tradeoff in between flexibility and consistency ultimately may be dependent on the setting. While cleaning answers for extraneous content may be consistent in the sorts of patterns encountered, there is a higher degree of variability in what candidate tails might be the `same'. For instance, a regex can consistently extract a variety of proper nouns, likely better than an LLM can extract an entity without finetuning; conversely the counter function cannot handle any instances of the same phrase with different punctuation while the LLM could recognize that both NYC and New York City are the `same'. 

\section{Conclusion}
Our work provides a preliminary investigation utilizing an LLM for generation, filtration, and parsing for the task of triple completion.
While this work focused on the addition of information, it is important to note that variations in prompt structure could yield better use of the information, as opposed to our usage which maintained a `promptless' approach for generating candidate tails wherever possible, with the exception of the information provided. Additionally, exploration of whether these findings are consistent across different model types and sizes could prove beneficial in future work. Our findings show, within our constrained setting, even without RAG or finetuning it is possible to improve the efficacy of LLM based AKBC. Both the introduction of additional LLM-provided information and LLM based filtration improved precision beyond simple LLM calls for completion. Additionally, we found LLM based consensus to be more effective than in-built python methods. However, we determined that a regex is more useful in handling the outputs despite the flexibility an LLM might provide. 

\section*{Declaration on Generative AI}
The authors have not employed any Generative AI tools in the preparation of this draft.

\bibliography{ref}

\end{document}